\documentclass[conference]{IEEEtran}
\IEEEoverridecommandlockouts

\usepackage{cite}
\usepackage{amsmath,amssymb,amsfonts}
\usepackage{graphicx}
\usepackage{textcomp}
\usepackage{xcolor}

\usepackage{booktabs}
\usepackage{multirow}
\usepackage{makecell}
\usepackage{algorithm}
\usepackage{algpseudocode}
\usepackage[colorlinks,linkcolor=blue]{hyperref}
\def\BibTeX{{\rm B\kern-.05em{\sc i\kern-.025em b}\kern-.08em
    T\kern-.1667em\lower.7ex\hbox{E}\kern-.125emX}}
\begin{document}

\title{Enhancing Visual Forced Alignment with Local Context-Aware Feature Extraction and Multi-Task Learning
\thanks{Shilin Wang is the corresponding author. This work was supported by the National Natural Science Foundation of China (62271307). Code is available at: \href{https://github.com/heyi0616/Visual-Forced-Alignment}{https://github.com/heyi0616/Visual-Forced-Alignment}.

© 2025 IEEE.  Personal use of this material is permitted.  Permission from IEEE must be obtained for all other uses, in any current or future media, including reprinting/republishing this material for advertising or promotional purposes, creating new collective works, for resale or redistribution to servers or lists, or reuse of any copyrighted component of this work in other works.}
}

\author{\IEEEauthorblockN{Yi He}
\IEEEauthorblockA{
\textit{Shanghai Jiao Tong University}\\
Shanghai, China \\
heyi96@sjtu.edu.cn}
\and
\IEEEauthorblockN{Lei Yang}
\IEEEauthorblockA{
\textit{Shanghai Jiao Tong University}\\
Shanghai, China \\
yangleisx@sjtu.edu.cn}
\and
\IEEEauthorblockN{Shilin Wang}
\IEEEauthorblockA{
\textit{Shanghai Jiao Tong University}\\
Shanghai, China \\
wsl@sjtu.edu.cn}

}

\maketitle

\begin{abstract}
    This paper introduces a novel approach to Visual Forced Alignment (VFA), aiming to accurately synchronize utterances with corresponding lip movements, without relying on audio cues. We propose a novel VFA approach that integrates a local context-aware feature extractor and employs multi-task learning to refine both global and local context features, enhancing sensitivity to subtle lip movements for precise word-level and phoneme-level alignment. Incorporating the improved Viterbi algorithm for post-processing, our method significantly reduces misalignments. Experimental results show our approach outperforms existing methods, achieving a 6\% accuracy improvement at the word-level and 27\% improvement at the phoneme-level in LRS2 dataset. These improvements offer new potential for applications in automatically subtitling TV shows or user-generated content platforms like TikTok and YouTube Shorts.
\end{abstract}

\begin{IEEEkeywords}
    visual forced alignment, multi-task learning, multi-modal application.
\end{IEEEkeywords}

\section{Introduction}
\label{sec:intro}

Visual Forced Alignment (VFA) is a technology that generates timeline information—specifically, the start and end times for each word or phoneme—in videos, based on corresponding textual content. VFA can be utilized to synchronize subtitles with utterances of speakers in films or TV shows, even in the absence of audio, ensuring consistency between lip movements and text. It can also assist Audio Forced Alignment (AFA)\cite{Moreno1998,Li2022,McAuliffe2017,Bain2023}, which aligns audio with text, particularly in noisy environments. As platforms like TikTok and YouTube Shorts encourage the upload of personal videos, the automated and accurate addition of subtitles to these videos merits further research. Moreover, precise word segmentation can aid researchers in analyzing speakers' talking habits, thereby enhancing the accuracy of speaker authentication\cite{Guo2023,Yang2020}.

The research on VFA remains less developed compared to AFA. Tasks such as Visual Speech Detection\cite{Prajwal2022,Roth2020} can identify speaking frames in a video but can not associate these frames with speech content. Other researchers\cite{Prajwal2021,Momeni2020} have explored Visual Keyword Spotting, which locates the duration of a single word. While repeating this process could align entire video, it risks causing overlap between adjacent words and demands substantial computational resources. The first dedicated VFA system, DVFA\cite{Kim2023}, predicts the category of each frame by integrating video and text features and incorporates an additional branch to evaluate the accuracy of the provided text.

\begin{figure*}[h]
    \centering
    \includegraphics[width=1\linewidth]{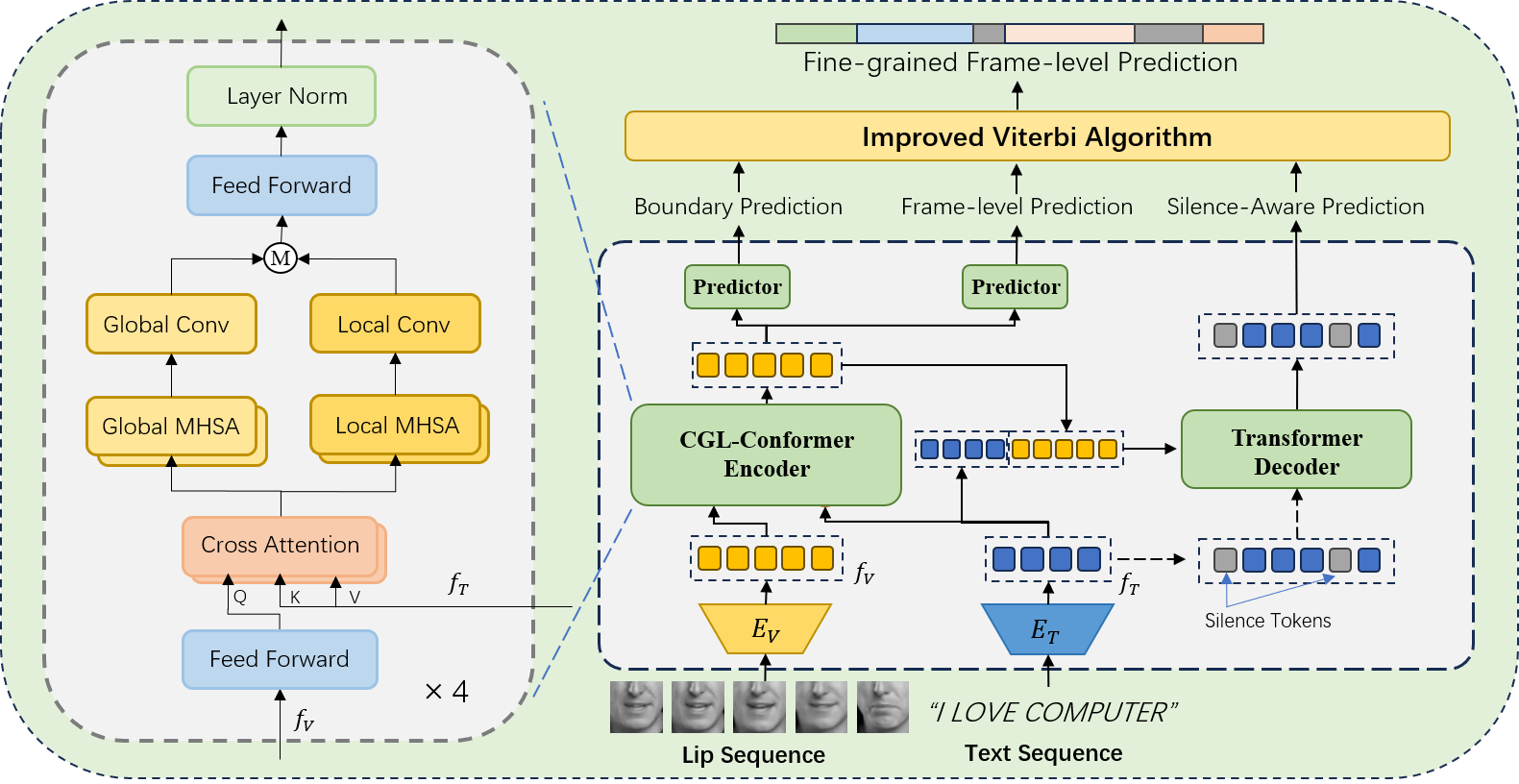}
    \caption{Architecture of the proposed method.}
    \label{fig:Arc}
\end{figure*}

Lip movements, which are subtler than vocal changes during transitions between words or phonemes, often go unnoticed and even may be overlooked due to video sampling intervals. This challenge is particularly significant in phoneme-level alignment, as phonemes are more fine-grained units than words (with each word typically comprising 2-4 phonemes). As a result, DVFA achieves only 61\% accuracy at the phoneme-level on the LRS2 dataset, compared to 84\% at the word-level. To improve alignment accuracy, it is necessary to rely on contextual semantic information and changes in lip shape to infer the timing of content switches. The former considers a longer range of temporal information, while the latter focuses on changes within a few frames. To address this challenge, we have designed a novel temporal feature extraction network that not only focuses on global context, but also refines local context features. This approach enhances the model's sensitivity to subtle changes in surrounding frames. The local context complements the global context, thereby aiding in finer-grained alignment.

In alignment tasks, such as word-level alignment, each frame may represent not only a spoken word but also a silent frame without articulation. Given a sequence of words, accurate segmentation of a speech video can theoretically be achieved by identifying the timing of content transitions—both between words and between words and silence—and detecting the presence of silence between words. To improve model accuracy in these areas, we developed a multi-task learning approach that includes frame-level, boundary, and silence-aware prediction. Furthermore, we noted that previous methods independently predicted the category of each frame, which could result in inconsistencies between the predicted sequence and the actual content. To address this, we employed the Viterbi algorithm \cite{Viterbi1967} for post-processing, integrating the outputs from multi-task learning to ensure precise forced alignment of the sequence. Our contributions are threefold:

\begin{itemize}
    \item We proposed the CGL-Conformer encoder, designed for Visual Forced Alignment tasks, which integrates video and textual information and pays attention to both global and local temporal information.
    \item We introduced a novel VFA network that utilizes multi-task learning and Viterbi algorithm to perform more detailed and accurate video segmentation.
    \item Our network achieved state-of-the-art results on the LRS2 and LRS3 datasets, particularly achieving a 28\% accuracy improvement at the phoneme-level in LRS2 dataset.
\end{itemize}

\section{The Proposed Method}
\label{sec:method}
The overall architecture of the proposed method is given in Fig.\ref{fig:Arc}. The model's input consists of a sequence of lip images from a speech video, along with the corresponding textual content. The lip sequence has a shape of $T_v\times H \times W \times 3$, where $T_v$ is the frame length, $H$ and $W$ are the height and width of the lip image. The text has a length of $T_t < T_V$ . We aim to determine the frames where each word or phoneme in the text begins and ends in the video. Video is embedded by a ResNet\cite{He2016} encoder with the first layer replaced by 3D convolution. Text tokens are embedded by an embedding layer. The channels are both projected to embedding size $C$.

\subsection{Cross Global-Local Conformer Encoder}
To adapt to the VFA task, we developed Cross Global-Local Conformer (CGL-Conformer), an enhanced Conformer\cite{Ma2021} network, as our cross-modal encoder. The encoder is shown in the left part of the Fig.\ref{fig:Arc}. Note that the residual connections for each block in the figure are omitted. First, a cross-attention layer is added before the self-attention layer to integrate video features and text features. The video features are passed in as queries, so the length of output feature is the same as the video length $T_v$.

After the information fusion, the encoder employs global multi-head self-attention\cite{Vaswani2017} to capture long-range dependencies, maintaining a comprehensive understanding of the input. However, excessive global information may cause local small changes in lip region to lose focus, which is an important basis for judging content changes. Therefore, local multi-head self-attention is introduced to refine local details and contextual nuances. Specifically, the attention mechanism at each time point is carried out within a window size $W$ centered around it, which can be described as:
\begin{equation}
    \text{Local Attention}(Q, K, V) = \text{softmax}\left(\frac{QK^TM}{\sqrt{C}}\right)V
\end{equation} 
$Q, K, V$ are the inputs of local attention. $M$ is the mask matrix where only diagonal elements within a predefined window around each are set to 1, with all other elements set to zero. 

The self-attention layers are followed by global and local convolution layers. In the original Conformer, depthwise separable convolutions\cite{Chollet2017} are used, including 1D pointwise and 1D depthwise convolution. The latter uses a large kernel of size 31. In order to pay more attention to local changes and align with local self-attention, we adopted convolutional kernels with smaller receptive fields in the local branch. The two branches are finally fused by Max Feature Map (MFM)\cite{Wu2018}, which selects the optimal feature by comparing the element-wise maximum between the global and local feature maps. This dual attention and convolution mechanism ensures a balance between broad context and fine details.

\subsection{Multi-Task Learning}
The proposed VFA model leverages multi-task learning to enhance its capacity to align text with silent video frames. This approach integrates three distinct but interconnected tasks to improve overall performance and robustness:

\noindent\textit{1) Frame-level prediction.} Each frame of the video is categorized into predefined classes using a cross-entropy loss function $\mathcal{L}_{\text{F}}$. The result of frame level prediction can be considered a preliminary alignment.

\noindent\textit{2) Boundary prediction.} The precise identification of boundaries where phonemes or words begin and end is crucial for temporal alignment. To address this, the model uses binary cross-entropy loss $\mathcal{L}_{\text{B}}$ to learn whether each frame represents a boundary. This task is specifically designed to improve the temporal resolution of the alignment, ensuring that the text is synced accurately with the visual cues.

\noindent\textit{3) Silence-aware text prediction.}  Considering the presence of silent frames, even if it is determined which frames belong to the boundary, there may still be a mismatch between the content and its boundary. Therefore, we designed a silence-aware text predictor. Specifically, the text embedding features and the output of the cross-modal encoder are concatenated and input into a Transformer decoder. Through this sequence to sequence structure and cross-entropy loss $\mathcal{L}_{\text{S}}$, a text sequence with silence tokens inserted is obtained. Together with boundary information, it can help the model further improve the accuracy of frame level segmentation.

The overall loss function is the summation of the aforementioned loss terms:
\begin{equation}
    \mathcal{L} = \mathcal{L}_{F} +  \mathcal{L}_{B} + \mathcal{L}_{S}
    \label{eq:Loss-total}
\end{equation}

\subsection{Viterbi Forced Alignment}
While simply selecting the predicted value with the highest probability of each frame can yield an alignment result, this approach may lead to discrepancies between the alignment and the order of the reference text, or even omit some tokens entirely. To address these issues, we employed the Viterbi algorithm, commonly used in audio forced alignment, to post-process the frame-level probabilities \textit{probs} and ensure they conform to the text sequence. The Viterbi algorithm utilizes dynamic programming to identify the most probable path that adheres to the prescribed order of the input text. We incorporated silence-aware predicted text \textit{sil\_aware\_label} as the input because the original text does not account for silence tokens, potentially leading to overlooked silence frames. Moreover, we integrated boundary probabilities \textit{bound\_pred} into the algorithm. If a frame is highly likely to be a boundary, we forcibly alter the text category of that frame. By combining three distinct predictions within the Viterbi algorithm, we not only enforce content alignment but also enhance alignment accuracy by emphasizing critical silence and boundary details. The comprehensive steps of the algorithm are delineated in Algorithm\ref{viterbi}.

\begin{algorithm}
    \caption{Improved Viterbi Algorithm}
    \label{viterbi}
    \begin{algorithmic}[1]
        \State \textbf{Input:} probs, sil\_aware\_label, bound\_pred
        \State \textbf{Output:} alignment
        \State $T \gets \text{len}(bound\_pred), C \gets \text{len}(sil\_aware\_label)$
        \State Initialize DP table \textit{dp} and path table \textit{path} with dimensions $(T+1) \times (C+1)$
        \State Set \textit{dp[0, :]} and \textit{dp[:, 0]} to $-\infty$, and \textit{dp[0, 0]} to $0$
        \For{$t = 1$ to $T$}
        \For{$c = 1$ to $C$}
        \State choices $\gets (dp[t-1, c-1], dp[t-1, c])$
        \If{$t \geq 2$ and bound\_pred[t-1] $> threshold$}
        \State path[t, c] $\gets$ 0  \Comment{Force transition to boundary}
        \Else
        \State path[t, c] $\gets$ \text{argmax}(choices) \Comment{Normal state}
        \EndIf
        \State dp[t, c] $\gets$ choices[path[t, c]] + $\log(\text{probs}[t-1, \text{sil\_aware\_label}[c-1]])$
        \EndFor
        \EndFor
        \State alignment $\gets$ empty list
        \State $t, c \gets T, C$
        \While{$t > 0$ and $c > 0$}
        \State alignment.append((t, c))
        \State $t, c \gets t - 1, c$ \textbf{if} path[t, c] $= 1$ \textbf{else} $t - 1, c - 1$
        \EndWhile
        \State alignment.reverse()
        \State \textbf{return} alignment
    \end{algorithmic}
\end{algorithm}

\section{Experiments}
\label{sec:exp}
\subsection{Datasets}
LRS2 dataset\cite{SonChung2017} and LRS3 dataset\cite{Afouras2018} are used to evaluate the performance of VFA. The LRS2 dataset is a comprehensive collection of naturally-occurring videos, featuring a variety of news and talk show clips sourced from the BBC. The LRS3 dataset includes TED and TEDx talks, offering a larger and more diverse set of spoken English sequences. The dataset splits both follow \cite{Kim2023}.

\subsection{Implementation Detail}
Montreal Forced Alignment\cite{McAuliffe2017}, an audio forced alignment tool, is used to generate the labels of alignment, following\cite{Kim2023}. Video frames are cropped to $96\times96$. Random horizontal flipping and time masking\cite{Ma2022} are both employed for data augmentation. 4-layer CGL-Conformer with embedding size of 512 is used. The window size $W$ for local attention is set to 16. The kernel of local convolution is set to 3. We use two layers of linear as boundary and frame-level predictor. The silence-aware decoder is a 2-layer transformer decoder. The threshold of boundary prediction in Viterbi algorithm is set to 0.8. The experiments were conducted on a single NVIDIA 2080 Ti GPU.

\subsection{Comparison with the State-of-the-Art}
To compare with other methods, we used the same experimental setup as DVFA. MAE (Mean Absolute Error) and ACC (Accuracy) are used as metrics. MAE is the time differences between predicted and true word/phoneme boundaries. ACC refers to the frame level prediction accuracy, which is obtained by dividing the number of correctly predicted frames by the total number of frames in a video. 

Our method is compared with four other methods, namely KWS-Net\cite{Momeni2020}, Transpotter\cite{Prajwal2021}, CTC-based method\cite{Kuerzinger2020}, and DVFA. The first two are methods for visual keyword spotting, while the third method utilizes CTC in the lip-reading model for alignment. DVFA is the first model specifically designed for VFA. The comparison results with them at the word-level are shown in Table \ref{tab:lrs2-word-compare}. It can be seen that our method outperforms SOTA on both LRS2 and LRS3 datasets, in terms of MAE and ACC metrics. This is because by focusing on more local temporal changes and adding emphasis to silence and boundary information, more detailed alignment results can be obtained.

In addition, we also conducted experiments at the phoneme-level, as shown in Table \ref{tab:lrs2-phoneme-compare}. Compared with DVFA, the improvement is more significant, with a 76\% decrease in MAE and a 27\% increase in ACC (in LRS2 dataset). Compared to words, phonemes are more fine-grained units. This further confirms that our method can focus on finer details and achieve more accurate frame level predictions.

\begin{table}[htbp]
    \caption{Visual forced alignment performance in word-level}
    \centering
    \resizebox{.98\columnwidth}{!}{
        \begin{tabular}{lcccccc}
            \toprule
            Dataset && \multicolumn{2}{c}{LRS2} && \multicolumn{2}{c}{LRS3} \\
            \cline{1-1} \cline{3-4} \cline{6-7} \noalign{\vskip -3pt}\\
            Model & &MAE \textdownarrow & ACC \textuparrow && MAE \textdownarrow & ACC \textuparrow\\
            \midrule
            KWS-Net\cite{Momeni2020} && 171.5ms & 53.0\% && 262.9ms & 42.6\%\\
            CTC-based\cite{Kuerzinger2020} && 80.4ms & 71.7\% && 124.5ms & 60.1\% \\
            Transpotter\cite{Prajwal2021} && 71.7ms & 75.0\% && 167.3ms & 61.8\% \\
            DVFA\cite{Kim2023} && 67.7ms & 84.2\% && 97.7ms & 80.2\%\\
            \midrule
            Proposed Method && \textbf{50.2ms} & \textbf{89.5\%} && \textbf{70.5ms} & \textbf{87.0\%}\\
            \bottomrule
        \end{tabular}
    }
    \label{tab:lrs2-word-compare}
\end{table}

\begin{table}[htbp]
    \caption{Visual forced alignment performance in phoneme-level}
    \centering
    \resizebox{.98\columnwidth}{!}{
        \begin{tabular}{lcccccc}
            \toprule
            Dataset && \multicolumn{2}{c}{LRS2} && \multicolumn{2}{c}{LRS3} \\
            \cline{1-1} \cline{3-4} \cline{6-7} \noalign{\vskip -3pt}\\
            Model & &MAE \textdownarrow & ACC \textuparrow && MAE \textdownarrow & ACC \textuparrow\\
            \midrule
            DVFA\cite{Kim2023} && 176.5ms & 61.3\% && 249.1ms & 57.1\%\\
            \midrule
            Proposed Method && \textbf{41.2ms} & \textbf{78.0\%} && \textbf{65.2ms} & \textbf{76.7\%}\\
            \bottomrule
        \end{tabular}
    }
    \label{tab:lrs2-phoneme-compare}
\end{table}

\begin{table}[htbp]
    \caption{Ablation study on LRS2 dataset in word-level}
    \centering
    \begin{tabular}{@{}lcc@{}}
        \toprule
        Model                           & MAE \textdownarrow & ACC \textuparrow   \\ \midrule
        baseline                        & 72.1ms & 86.9\% \\
        \hspace{3mm}+ local branch     & 71.2ms & 87.8\%\\
        \hspace{6mm}+ multi-task learning     & 68.2ms & 88.0\% \\
        \hspace{9mm}+ improved Viterbi  & 50.2ms & 89.5\% \\ \bottomrule
    \end{tabular}
    \label{tab:lrs2-abalation}
\end{table}

\subsection{Ablation Study}
\begin{figure}[h]
    \centering
    \includegraphics[width=1\linewidth]{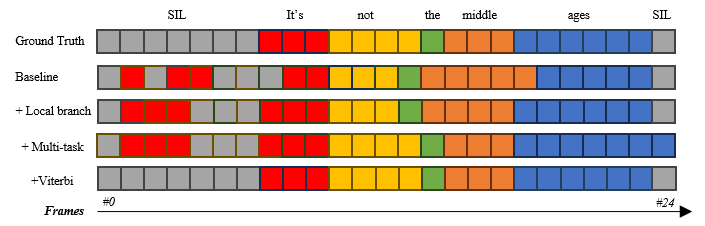}
    \caption{Visualization of the alignment result.}
    \label{fig:Ablation}
\end{figure}

We conducted an ablation study, as shown in Table \ref{tab:lrs2-abalation}, to evaluate the effectiveness of each module. The alignment was performed at the word-level using the LRS2 dataset. The baseline model employs the original Conformer and relies solely on frame-level prediction loss. The results show that the addition of each component, including the local branch and multi-task learning, enhances model performance. Notably, the integration of the improved Viterbi algorithm significantly improves alignment outcomes. This improvement is attributed to the comprehensive use of textual information, which reduces the possibility of disordered results, and the integration of outputs from multi-task learning, which refines content boundaries. We visualized the frame-level prediction results after incorporating these modules in Fig. \ref{fig:Ablation}, where different colors denote different classes. The label \textit{SIL} indicates the absence of sound. The visualization demonstrates that both the local branch and multi-task learning progressively enhance the accuracy of predicted boundaries. Following post-processing with the improved Viterbi algorithm, precise forced alignment between the predicted sequences and the text was achieved.

\section{Conclusion}
\label{sec:conclusion}
In conclusion, our enhanced visual forced alignment model leverages the improved Viterbi algorithm along with silence-aware, boundary, and frame-level predictions to substantially improve alignment accuracy. This integration ensures precise text synchronization with visual data. Experiments on LRS2 and LRS3 datasets demonstrate SOTA performance of our model, marking a significant advancement in visual forced alignment.


\vfill\pagebreak

{\small
\bibliographystyle{IEEEtran}
\bibliography{refs}
}

\end{document}